\documentclass{article}

\usepackage[preprint,nonatbib]{nips_2018}

\usepackage[utf8]{inputenc} % allow utf-8 input
\usepackage[T1]{fontenc}    % use 8-bit T1 fonts
\usepackage{hyperref}       % hyperlinks
\usepackage{url}            % simple URL typesetting
\usepackage{booktabs}       % professional-quality tables
\usepackage{amsfonts}       % blackboard math symbols
\usepackage{nicefrac}       % compact symbols for 1/2, etc.
\usepackage{microtype}      % microtypography
\usepackage{listings}
\usepackage{ascmac}
\usepackage[utf8]{inputenc}
\usepackage{graphicx}
\usepackage{wrapfig}
\usepackage{amsmath,amsfonts,amssymb,amsthm}

\usepackage{color}
\usepackage{soul}  % For highlighting (requires color package)

\begin{document}

\title{End-to-End Audio Visual Scene-Aware Dialog 
using Multimodal Attention-Based Video Features}
\author{
%--------------------
  Chiori Hori$^{\dagger}$,
  Huda Alamri$^{*\dagger}$,
  Jue Wang$^{\dagger}$,
  Gordon Wichern$^{\dagger}$,\\  
\bf Takaaki Hori$^{\dagger}$,  
  Anoop Cherian$^{\dagger}$,
  Tim K. Marks$^{\dagger}$,  \\
\bf Vincent Cartillier$^{*}$,
  Raphael Gontijo Lopes$^{*}$,
  Abhishek Das$^{*}$, \\
\bf Irfan Essa$^{*}$,
  Dhruv Batra$^{*}$
  Devi Parikh$^{*}$, 
  \\[8pt]
%--------------------
$^{\dagger}$Mitsubishi Electric Research Laboratories (MERL), Cambridge, MA, USA \\
%\texttt{\{juewang, cherian, tmarks, chori\}@merl.com}\\
%--------------------
%-------------------
$^{*}$School of Interactive Computing, Georgia Tech\\
%\texttt{\{halamri, vcartillier3, rgl3, ifran, parikh, batra\}@gatech.edu}\\ 
%--------------------
}

\maketitle
\begin{abstract}
  Dialog systems need to understand dynamic visual scenes in order
  to have conversations with users about the objects and events around them. 
  Scene-aware dialog systems for real-world applications could be developed 
  by integrating state-of-the-art technologies
  from multiple research areas, including: end-to-end dialog technologies, 
  which generate system responses using models trained from dialog data; 
  visual question answering (VQA) technologies, which answer 
  questions about images using learned image features; and
  video description technologies, in which descriptions/captions are generated from videos using multimodal information.
  We introduce a new dataset of dialogs about videos of human behaviors.
  Each dialog is a typed conversation that consists of a sequence of 10 question-and-answer (QA) pairs between two Amazon Mechanical Turk (AMT) workers.
  In total, we collected dialogs on $\sim9,000$ videos.
  Using this new dataset, we trained an end-to-end conversation model that generates responses in a dialog about a video.
  Our experiments demonstrate that using multimodal features that were developed for multimodal attention-based video description enhances the quality of generated dialog about dynamic scenes (videos).
  Our dataset, model code and pretrained models will be publicly available 
  for a new Video Scene-Aware Dialog challenge.
\end{abstract}

\section{Introduction}
Spoken dialog technologies have been applied 
in real-world human-machine interfaces including smart phone digital assistants, 
car navigation systems, voice-controlled smart speakers, and  human-facing robots \cite{mctear2002spoken,young2000probabilistic, zue2000juplter}. 
Generally, a dialog system consists of a pipeline of data processing modules, 
including automatic speech recognition, spoken language understanding, 
dialog management, sentence generation, and speech synthesis.
However, all of these modules require significant hand engineering and domain knowledge for training.
Recently, end-to-end dialog systems have been gathering attention,
and they obviate this need for expensive hand engineering to some extent.
In end-to-end approaches, dialog models are trained using only paired input and output sentences, without relying on pre-designed data processing modules or intermediate internal data representations such as concept tags and slot-value pairs. 
End-to-end systems can be trained to directly map from a user's utterance to a system response sentence and/or action. This significantly reduces the data preparation and system development cost.
Several types of sequence-to-sequence models have been applied to end-to-end dialog systems, and it has been shown that they can be trained in a completely data-driven manner.
End-to-end approaches have also been shown to better handle flexible conversations between
the user and the system by training the model on large conversational datasets \cite{vinyals2015neural,lowe2015ubuntu}.

In these applications, however, all conversation is triggered by user speech input, 
and the contents of system responses are limited by the training data (a set of dialogs). 
Current dialog systems cannot understand dynamic scenes using multimodal sensor-based input

such as vision and non-speech audio, so machines using such dialog systems cannot 
have a conversation about what's going on in their surroundings.
To develop machines that can carry on a conversation about objects and events 
taking place around the machines or the users, dynamic scene-aware dialog technology is essential.

To interact with humans about visual information, 
systems need to understand both visual scenes and natural language inputs.
One naive approach could be a pipeline system in which the output of 
a visual description system is used as an input to a dialog system. In this cascaded approach, semantic frames such as "who" is doing "what" and "where"
must be extracted from the video description results. The prediction of frame type and 
the value of the frame must be trained using annotated data.
In contrast, the recent revolution of neural network models 
allows us to combine different modules into a single end-to-end differentiable network.
We can simultaneously input video features and user utterances 
into an encoder-decoder-based system 
whose outputs are natural-language responses.

Using this end-to-end framework, 
{\em visual question answering} (VQA) has been intensively researched 
in the field of computer vision \cite{VQA,balanced_binary_vqa,balanced_vqa_v2}.
The goal of VQA is to generate answers to questions about an imaged scene, 
using the information present in a single static image.
As a further step towards conversational visual AI, 
the new task of {\em visual dialog} was introduced~\cite{DBLP:journals/corr/DasKGSYMPB16}, 
in which an AI agent holds a meaningful dialog with humans about an image using natural, conversational language \cite{visdial_rl}.
%, as shown in the example in Figure~\ref{fig:vd}. 
While VQA and visual dialog take significant steps towards human-machine interaction, 
they only consider a single static image.
To capture the semantics of dynamic scenes, recent research has focused on {\em video description} (natural-language descriptions of videos). The state of the art in video description uses a multimodal 
attention mechanism that selectively attends to different input modalities (feature types), such spatiotemporal motion features and audio features, in addition to temporal attention~\cite{Hori_2017_ICCV}. 

In this paper, we propose a new research target, a dialog system that can discuss dynamic scenes with humans, which lies at the intersection of multiple avenues of research in natural language processing, computer vision, and audio processing. To advance this goal, we introduce a new model that incorporates technologies for multimodal attention-based video description into an end-to-end dialog system. We also introduce a new dataset of human dialogues about videos.  
We are making our dataset, code, and model publicly available for a new Video Scene-Aware Dialog Challenge. 

%%%%%%%%%%%%%%%%%%%%%%%%%%%%%%%%%%%%%%%%%%%%%%%%%%%%%%%
\section{Audio Visual Scene-Aware Dialog Dataset}
%%%%%%%%%%%%%%%%%%%%%%%%%%%%%%%%%%%%%%%%%%%%%%%%%%%%%%%
\label{sec:video-features}
We collected  text-based conversations data about short videos for Audio Visual Scene-Aware Dialog (AVSD) as described in \cite{alamri2018audio} using from an existing video description dataset,  
Charades~\cite{sigurdsson2016hollywood}, for Dialog System Technology Challenge the 7th edition (DSTC7)\footnote{http://workshop.colips.org/dstc7/call.html}. 
Charades is an untrimmed and multi-action dataset, 
containing 11,848 videos split into 7985 for training, 
1863 for validation, and 2,000 for testing. 
It has 157 action categories, with several fine-grained actions. 
Further, this dataset also provides 27,847 textual descriptions for the videos, 
each video is associated with 1--3 sentences. 
As these textual descriptions are only available in the training and validation set, 
we report evaluation results on the validation set. 

The data collection paradigm for dialogs was similar to the one described in~\cite{DBLP:journals/corr/DasKGSYMPB16}, 
in which for each image, two different Mechanical Turk workers interacted via a text interface to yield a dialog. In~\cite{DBLP:journals/corr/DasKGSYMPB16}, 
each dialog consisted of a sequence of questions and answers about an image.
In the video scene-aware dialog case, 
two Amazon Mechanical Turk (AMT) workers had a discussion about events in a video.
One of the workers played the role of an answerer who had already watched the video. 
The answerer answered questions asked by another AMT worker -- the questioner.
The questioner was not allowed to watch the whole video but 
only the first, middle and last frames of the video which were single static images.
After having a conversation to to ask about the events that happened between the frames 
through 10 rounds of QA,
the questioner summarized the events in the video as a description.

In total, we collected dialogs for 7043 videos from the Charades training set and 
all of the validation set (1863 videos).
Since we did not have scripts for the test set,
we split the validation set into 732 and 733 videos 
and used them as our validation and test sets respectively.
See Table~\ref{tab:data} for statistics.
The average numbers of words per question and answer are 8 and 10, respectively.

%%-------------------------
\begin{table}[h]
\centering
\caption{Video Scene-aware Dialog Dataset on Charades}
\label{tab:data}
\begin{tabular}{ll|ccc}
\hline
& 
& training 
& validation 
& test \\
\hline
\#dialogs 
& 
& 6,172 %7043
& 732  
& 733 \\
\#turns   
& 
& 123,480
%123,440 (?)  
& 14,680
%14,640(?)  
& 14,660
%14,660(?) 
\\
\#words   
& 
&  1,163,969
&  138,314
&  138,790
\\
\hline
\end{tabular}
\end{table}
%%-------------------------

%%%%%%%%%%%%%%%%%%%%%%%%%%%%%%%%%%%%%%%%%%%%%%%%%%%%
\section{Video Scene-aware Dialog System}
%%%%%%%%%%%%%%%%%%%%%%%%%%%%%%%%%%%%%%%%%%%%%%%%%%%%

\begin{figure}[bt]
	\centering
	\centerline{\includegraphics[width=13.5cm]{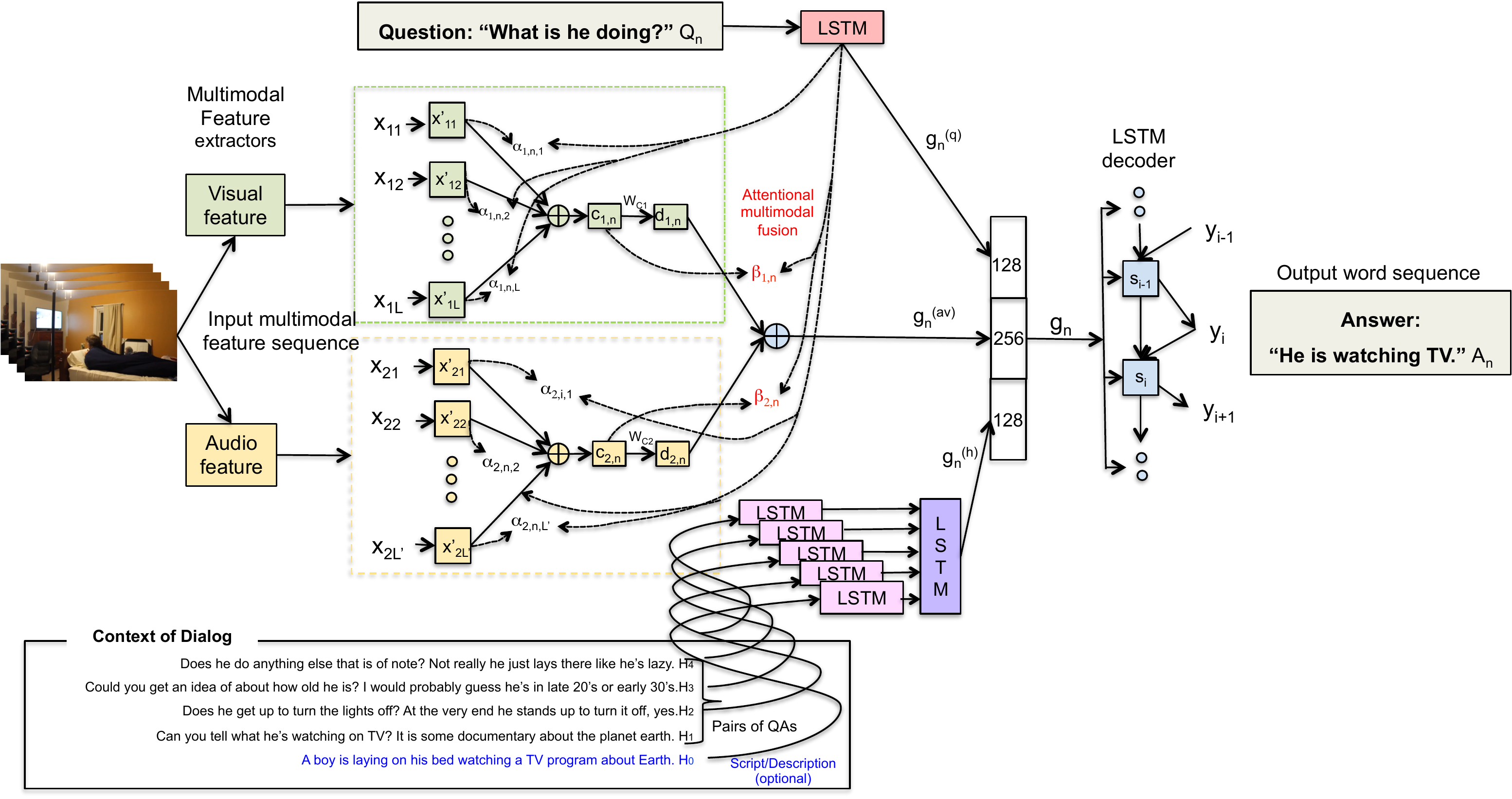}}
	\caption{Our multimodal-attention based video scene-aware dialog system}
	\label{fig:scene-aware-dialog}
	\vspace{-3mm}
\end{figure}

We built an end-to-end dialog system that can generate answers 
in response to user questions about events in a video sequence.
Our architecture is similar to the Hierarchical Recurrent Encoder in Das~\textit{et al.}~\cite{DBLP:journals/corr/DasKGSYMPB16}.
The question, visual features, and the dialog history are fed into corresponding LSTM-based encoders
to build up a context embedding, 
and then the outputs of the encoders are fed into a LSTM-based decoder to generate an answer.
The history consists of encodings of QA pairs. 
We feed multimodal attention-based video features into the LSTM encoder instead of single static image features.
Figure \ref{fig:scene-aware-dialog} shows the architecture of our video scene-aware dialog system.

%--- Fig.1 at Page 3--------------------------------------------
%---------------------------------------------------------------

%%%%%%%%%%%%%%%%%%%%%%%%%%%%%%%%%%%%%%%%%%%%%%%%%
\subsection{End-to-end Conversation Modeling}
%%%%%%%%%%%%%%%%%%%%%%%%%%%%%%%%%%%%%%%%%%%%%%%%%
\label{sec:neural-conversation}
% Can we just use tne name of the model here instead of "a well known ..."?
This section explains the neural conversation model of \cite{vinyals2015neural}, which is designed as a sequence-to-sequence mapping process using recurrent neural networks (RNNs). Let $X$ and $Y$ be input and output sequences, respectively.
The model is used to compute posterior probability distribution $P(Y|X)$.
For conversation modeling, $X$ corresponds to the sequence of previous sentences in a conversation, and $Y$ is the system response sentence we want to generate.  In our model, both $X$ and $Y$ are sequences of words.  $X$ contains all of the previous turns of the conversation, concatenated in sequence, separated by markers that indicate to the model not only that a new turn has started, but which speaker said that sentence. 
The most likely hypothesis of $Y$ is obtained as
\begin{align}
\hat{Y}&=\arg \max_{Y\in {\cal V}^*} P(Y|X) \label{eq:generate}\\
    &=\arg \max_{Y\in {\cal V}^*} \prod_{m=1}^{|Y|} P(y_m|y_1,\dots,y_{m-1},X),
\end{align}
where ${\cal V}^*$ denotes a set of sequences of zero or more words in system vocabulary $\cal V$.
%Figure \ref{fig:lstm} illustrates a sequence-to-sequence model using RNNs. The previous sentences $X$ is fed to the encoder network (left-hand side) and then the hidden activation vector is fed to the decoder network (right-hand side).

Let $X$ be word sequence $x_1,\dots,x_T$ and $Y$ be word sequence $y_1,\dots,y_M$. The encoder network is used to obtain hidden states $h_t$ for $t=1,\dots,T$ as:
\begin{align}
h_t& = \mbox{LSTM}\left(x_t, h_{t-1}; \theta_{enc}\right),
\end{align}
where $h_0$ is initialized with a zero vector. $\mbox{LSTM}(\cdot)$ is a LSTM function with parameter set $\theta_{enc}$.
 
The decoder network is used to compute probabilities $P(y_m|y_1,\dots,y_{m-1},X)$ for $m=1,\dots,M$ as:
\begin{align}
s_0& = h_T \label{eq:decinit}\\
s_m& = \mbox{LSTM}\left(y_{m-1}, s_{m-1}; \theta_{dec}\right) \label{eq:declstm}\\
P({\bf y}|y_1,\dots,y_{m-1},X)&=\mbox{softmax}(W_o s_m + b_o),\label{eq:decpred}
\end{align}

where $y_0$ is set to {\tt<eos>}, a special symbol representing the end of sequence. $s_m$ is the $m$-th decoder state. $\theta_{dec}$ is a set of decoder parameters, and $W_o$ and $b_o$ are a matrix and a vector. 
In this model, the initial decoder state $s_0$ is given by the final encoder state $h_T$ as in Eq. \eqref{eq:decinit}, and the probability is estimated from each state $s_m$.
To efficiently find $\hat{Y}$ in Eq. \eqref{eq:generate}, we use a beam search technique since it is computationally intractable to consider all possible $Y$.

In the scene-aware-dialog scenario, a scene context vector including audio and visual features is also fed to the decoder. We modify the LSTM in Eqs. \eqref{eq:decinit}--\eqref{eq:decpred} as
\begin{align}
s_{n,0}& = \bar{\bf 0}\\
s_{n,m}& = \mbox{LSTM}\left([y_{n,m-1}^\intercal, g_n^\intercal]^\intercal, s_{n,m-1}; \theta_{dec}\right),\\
P({\bf y_n}|y_{n,1},\dots,y_{n,m-1},X)&=\mbox{softmax}(W_o s_{n,m} + b_o),
\end{align}
where $g_n$ is the concatenation of question encoding $g_n^{(q)}$, audio-visual encoding $g_n^{(av)}$ and history encoding $g_n^{(h)}$ for generating the $n$-th answer $A_n=y_{n,1},\dots,y_{n,|Y_n|}$. Note that unlike Eq. \eqref{eq:decinit}, we feed all contextual information to the LSTM at every prediction step. This architecture is more flexible since the dimensions of encoder and decoder states can be different.

$g_n^{(q)}$ is encoded by another LSTM for the $n$-th question, and $g_n^{(h)}$ is encoded with hierarchical LSTMs, where one LSTM encodes each question-answer pair and then the other LSTM summarizes the question-answer encodings into $g_n^{(h)}$.
The audio-visual encoding is obtained by multi-modal attention described in the next section.

\if 0
%--- Fig.1 -------------------------------------------------
\begin{figure}[t]
	\centering
	\centerline{\includegraphics[width=7.8cm]{system-architecture.pdf}}
	\caption{Scene aware dialog system}
	\label{fig:system-arch}
	\vskip -2mm
\end{figure}
%-----------------------------------------------------------
%--- Fig.2 -------------------------------------------------
\begin{figure}[t]
    \begin{minipage}{1.0\hsize}
	    \centering
	    \centerline{\includegraphics[width=7.5cm]{LSTM.pdf}}
	    \caption*{(a) LSTM-based encoder decoder \cite{vinyals2015neural}}
    \end{minipage}
    \vskip 2mm
    \begin{minipage}{1.0\hsize}
        \centering
	    \centerline{\includegraphics[width=7.5cm]{BLSTM.pdf}}
	    \caption*{(b) BLSTM-based encoder decoder}
	\end{minipage}
	%\vskip 5mm	
    \begin{minipage}{1.0\hsize}
	    \centering
	    \centerline{\includegraphics[width=7.5cm]{HRED.pdf}}
	    \caption*{(c) Hierarchical recurrent encoder decoder (HRED) \cite{serban2016building}}
	\end{minipage}
    \vskip -2mm
	\caption{Sequence-to-sequence models}
	\label{fig:models}
	\vskip -5mm
\end{figure}
%-----------------------------------------------------------
\fi

%%%%%%%%%%%%%%%%%%%%%%%%%%%%%%%%%%%%%%%%%%%%%%%
\subsection{Multimodal-attention based Video Features}
%%%%%%%%%%%%%%%%%%%%%%%%%%%%%%%%%%%%%%%%%%%%%%%
To predict a word sequence in video description,
prior work~\cite{attention4spatial@CVPR2016} extracted content vectors from image features of VGG-16 and 
spatiotemporal motion features of C3D, 
and combined them into one vector in the fusion layer as:
\begin{equation}
\label{eqn:fusion}
g_n^{(av)} = \tanh\left(\sum_{k=1}^K d_{k,n}\right),
\end{equation}
where
\begin{equation}
d_{k,n} = W_{ck}^{({\lambda}_D)} c_{k,n} + b_{ck}^{({\lambda}_D)},
\end{equation}
and $c_{k,n}$ is a context vector obtained using the $k$-th input modality.

%Figure \ref{fig:feature-fusion} illustrates this approach, which 
We call this approach Na\"ive Fusion, in which
multimodal feature vectors are combined using projection matrices $W_{ck}$ for $K$ different modalities (input sequences $x_{k1},\dots,x_{kL}$ for $k=1,\dots,K$).

\if 0
However, these feature vectors are combined in the sentence generation step with projection matrices $W_{c1}$ and $W_{c2}$, which do not depend on time. 
Consequently, for each modality (or each feature type), all of the feature vectors from that modality are given the same weight during fusion, independent of the decoder state. 
Note that Na\"ive Fusion is a type of late fusion, because the inherent difference in sampling rate of the three feature streams precludes early fusion (concatenation of input features).
The Na\"ive Fusion architecture lacks the ability to exploit multiple types of features effectively, because it does not allow the relative weights of each modality (of each feature type) to change based on the context of each word in the sentence.
\fi 

To fuse multimodal information, prior work~\cite{Hori_2017_ICCV} proposed method extends the attention mechanism.
We call this fusion approach {\em multimodal attention}. 
The approach can pay attention to specific modalities of input based on the current state of the decoder
to predict the word sequence in video description.
The number of modalities indicating the number of sequences of input feature vectors is denoted by
$K$.

The following equation shows an approach to perform the attention-based feature fusion:
\begin{equation}
g_n^{(av)} = \tanh\left(\sum_{k=1}^K {\beta}_{k,n} d_{k,n}\right).
\end{equation}

The similar mechanism for temporal attention is applied 
to obtain the multimodal attention weights ${\beta}_{k,n}$:
\begin{equation}
\beta_{k,n}=\frac{\exp(v_{k,n})}{\sum_{\kappa=1}^{K} \exp(v_{\kappa,n})},
\label{eqn:alpha}
\end{equation}
where
\begin{equation}
v_{k,n}=w^{\intercal}_B \tanh(W_B g_n^{(q)} + V_{Bk} c_{k,n} + b_{Bk}).
\end{equation}
Here the multimodal attention weights are determined by question encoding $g_n^{(q)}$ and the context vector of each modality $c_{k,n}$ as well as temporal attention weights in each modality.
$W_B$ and $V_{Bk}$ are matrices, $w_B$ and $b_{Bk}$ are vectors, and $v_{k,n}$ is a scalar.
%Figure~\ref{fig:feature-attention} shows the architecture of our sentence generator, including the multimodal attention mechanism. 
The multimodal attention weights 
can change according to the question encoding and the feature vectors 
(shown in Figure~\ref{fig:scene-aware-dialog}). 

This enables the decoder network to attend to a different set of features and/or modalities when predicting each subsequent word in the description.
Na\"ive fusion can be considered a special case of Attentional fusion,
in which all modality attention weights, $\beta_{k,n}$, are constantly 1. 

%``na\"ive Fusion (V)''
%%%%%%%%%%%%%%%%%%%%%%%%%%%%%%%%%%%%%%%%%%%%%%%
\section{Experiments for Multimodal attention-based Video Features}
%%%%%%%%%%%%%%%%%%%%%%%%%%%%%%%%%%%%%%%%%%%%%%%
To select best video features for the video scene-aware dialog system, 
we firstly evaluate the performance of video description
using multimodal attention-based video features in this paper.

%%%%%%%%%%%%%%%%%%%%%%%%%%%%%%%%%%%%%%%%%%%%%%%
\subsection{Datasets}
%%%%%%%%%%%%%%%%%%%%%%%%%%%%%%%%%%%%%%%%%%%%%%%
We evaluated our proposed feature fusion using 
the MSVD (YouTube2Text) \cite{guadarrama2013youtube2text}, 
MSR-VTT \cite{MSR-VTT:CVPR16}, and  Charades \cite{sigurdsson2016hollywood} video data sets.

\begin{itemize}
\item {MSVD (YouTube2Text) covers a wide range of topics including sports, animals, and music. 
We applied the same condition defined by \cite{guadarrama2013youtube2text}: 
a training set of 1,200 video clips, a validation set of 100 clips, and a test set of the remaining 670 clips.} \\
\item  {MSR-VTT is split into training, validation, and testing sets of 6,513, 497, and 2,990 clips respectively.
However, approvimatebly 12\@ of the MSR-VTT videos on YouTube have been removed.
We used the available data consists of 5,763, 419, and 2,616 clips for train, validation, and test respectively
defined by \cite{Hori_2017_ICCV}.}\\
\item {Charades~\cite{sigurdsson2016hollywood} is split into 7985 clips for training and 1863 clips for validation.
provides 27,847 textual descriptions for the videos, 
%each video is associated with 1--3 sentences. 
As these textual descriptions are only available in the training and validation set, 
we report the evaluation results on the validation set.} \\
\end{itemize}
Details of textual descriptions are summarized in Table \ref{tab:description-info}.
%27,847 video descriptions,
%66,500 temporally localized intervals for 157 action classes and
%41,104 labels for 46 object classes

%---------------------------------------------------------
\begin{table*}[ht]
%\begin{center}
\footnotesize
\centering
\caption{Sizes of textual descriptions in MSVD (YouTube2Text), MSR-VTT and Charades}
\label{tab:description-info}
\begin{tabular}{l|r|r|r|r|r}
\hline
& & & \multicolumn{1}{|c|}{\#Descriptions} &  & 
\multicolumn{1}{|c}{Vocabulary} \\
\raisebox{2mm}[0ex]{Dataset} & \raisebox{2mm}[0ex]{\#Clips} & \raisebox{2mm}[0ex]{\#Description} & \multicolumn{1}{|c|}{per clip} & \raisebox{2mm}[0ex]{\#Word} & 
\multicolumn{1}{|c}{size} \\
\hline
MSVD     & 1,970 & 80,839~ & 41.00 & 8.00 & 13,010   \\
MSR-VTT  & 10,000 & 200,000~ & 20.00 & 9.28 & 29,322  \\
Charades & 9,848   & 16,140~ & 1.64 & 13.04  & 2,582  \\
\hline
\end{tabular}
%\end{center}
%\vskip -3mm
\end{table*}
%----------------------------------------------------------

%%%%%%%%%%%%%%%%%%%%%%%%%%%%%%%%%%%%%%%%%%%%%%%
\subsection{Video Processing}
%%%%%%%%%%%%%%%%%%%%%%%%%%%%%%%%%%%%%%%%%%%%%%%
We used a sequence of 4096-dimensional feature vectors of
the output from the fully-connected fc7 layer of a VGG-16 network %~\cite{VggNet} 
pretrained on the ImageNet dataset for the image features.

The pretrained C3D~\cite{C3D} model is used 
to generate features for model motion and short-term spatiotemporal activity.
The C3D network reads sequential frames in the video and outputs a fixed-length feature vector every 16 frames. 
4096-dimensional features of activation vectors from fully-connected fc6-1 layer was applied to spatiotemporal features.

In addition to the VGG-16 and C3D features, we also adopted the state-of-the-art I3D  features~\cite{carreira2017quo}, spatiotemporal features that were developed for action recognition. The I3D model inflates the 
%regular 
2D filters and pooling kernels in the Inception V3 network 
along their temporal dimension, building 3D spatiotemporal ones. 
% Move this description to Discussion:
%In comparison to C3D features, that use the VGG-16 base architecture, I3D uses a more powerful Inception-V3 network architecture and has been pre-trained on the larger (and cleaner) Kinectics~\cite{kay2017kinetics} dataset. As a result, it has demonstrated state-of-the-art performances for the task of human action recognition in video sequences~\cite{carreira2017quo}. The Inception-V3 architecture has significantly less number of network parameters in comparison to the VGG-16 network. As a result, the I3D model can take as input longer sequences (64 frames) as against 16 frames in C3D -- such longer temporal receptive fields lead to better characterization of the spatiotemporal action dynamics. 
% How to extract features using I3D
%The I3D model takes as input a set of consecutive frames (16), each frame is of size 224x224, and generates a 2x7x7x1024 dimensional feature tensor as output of the 'Max5c' layer. We reshape this tensor to 2x1024 by averaging over the 7x7 spatial dimensions, which when reshaped produces the 1x2048 dimensional vector. This vector is what we use in the pkl file. We generate such 1x2048 features over the full sequences using a sliding window with a temporal stride of 16.
We used the output from the "Mixed\_5c" layer of the I3D network to be used as video features in our framework. 
As a pre-processing step, we normalized all the video features to have zero mean and unit norm; 
the mean was computed over all the sequences in the training set for the respective feature.

In the experiments in this paper, we treated I3D-rgb (I3D features computed on a stack of 16 video frame images) and I3D-flow (I3D features computed on a stack of 16 frames of optical flow fields) as two separate modalities that are input to our multimodal attention model. To emphasize this, we refer to I3D in the results tables as I3D (rgb-flow).

%%%%%%%%%%%%%%%%%%%%%%%%%%%%%%%%%%%%%%%%%%%%%%%
\subsection{Audio Processing}
%%%%%%%%%%%%%%%%%%%%%%%%%%%%%%%%%%%%%%%%%%%%%%%
While the original MSVD (YouTube2Text) dataset does not contain audio features, we were able to collect audio data for 1,649 video clips (84\% of the dataset) from the video URLs. In our previous work on multimodal attention for video description, we used two different types of audio features: concatenated mel-frequency cepstral coefficient (MFCC) features~\cite{hori2017attention}, and SoundNet~\cite{aytar2016soundnet} features~\cite{hori2017early}.  In this paper, we also evaluate features extracted using a new state-of-the-art model, Audio Set VGGish~\cite{hershey2017VGGish}.

Inspired by the VGG image classification architecture
%~\cite{VggNet}: This paper is cited by VGGish
(Configuration A without the last group of convolutional/pooling layers), the Audio Set VGGish model operates on 0.96 s log Mel spectrogram patches extracted from 16 kHz audio, and outputs a 128-dimensional embedding vector.  The model was  trained to predict an ontology of labels from only the audio tracks of millions of YouTube videos. In this work, we overlap frames of input to the VGGish network by 50\%, meaning an Audio Set VGGish feature vector is output every 0.48 s.  For SoundNet~\cite{aytar2016soundnet}, in which a fully convolutional architecture was trained to predict scenes and objects using a pretrained image model as a teacher, we take as input to the audio encoder the output of the second-to-last convolutional layer, which gives a 1024-dimensional feature vector every 0.67 s, and has a receptive field of approximately 4.16 s. For raw MFCC features, sequences of 13-dimensional MFCC features are extracted from 50 ms windows, every 25 ms, and then 20 consecutive frames are concatenated into a 260-dimensional vector and normalized to zero mean/unit variance (computed over the training set) and used as input to the BLSTM audio encoder.  

%%%%%%%%%%%%%%%%%%%%%%%%%%%%%%%%%%%%%%%%%%%%%%%
\subsection{Experimental Setup}
The caption generation model, i.e., the decoder network, is trained to minimize the cross entropy criterion using the training set.
Image features and deep audio features (SoundNet and VGGish) are fed to the decoder network through one projection layer of 512 units, while MFCC audio features are fed to a BLSTM encoder (one projection layer of 512 units and bidirectional LSTM layers of 512 cells) followed by the decoder network. The decoder network has one LSTM layer with 512 cells.
Each word is embedded to a 256-dimensional vector when it is fed to the LSTM layer.
In this video description task, 
we used L2 regularization for all experimental conditions and used RMSprop optimization.
%The LSTM and attention models were implemented using Chainer~\cite{tokui2015chainer}.

%%%%%%%%%%%%%%%%%%%%%%%%%%%%%%%%%%%%
\subsection{Evaluation}
%%%%%%%%%%%%%%%%%%%%%%%%%%%%%%%%%%%%
 The quality of the automatically generated sentences will be evaluated with objective measures 
 to measure the similarity between the generated sentences and ground truth sentences. %(see Figure \ref{fig:task}).
\if 0
%-------- Figure 5 ---------------------------
\begin{figure*}[h]
\centering
\centerline{\includegraphics[width=12.0cm]{example.png}}
\caption{Sentence generation and evaluation in the end-to-end conversation modeling track.}
\label{fig:task}
\end{figure*}
%-------------------------------------------
\fi
We will use the evaluation code for MS COCO caption generation\footnote{\url{https://github.com/tylin/coco-caption}} for objective evaluation of system outputs, which is a publicly available tool supporting various automated metrics for natural language generation such as BLEU, METEOR, ROUGE\_L, and CIDEr.

%%%%%%%%%%%%%%%%%%%%%%%%%%%%%%%%%%%%%%%%%%%%%%%%%%%
\subsection{Results and Discussion}
%%%%%%%%%%%%%%%%%%%%%%%%%%%%%%%%%%%%%%%%%%%%%%%%%%%
Tables \ref{table:result1}, \ref{table:result2}, and~\ref{table:result3} show the evaluation results on the MSVD (YouTube2Text), MSR-VTT Subset, and Charades datasets.
%I3D
The I3D spatiotemporal features outperformed the combination of VGG-16 image features and C3D spatiotemporal features. We also tried a combination of VGG-16 image features plus I3D spatiotemporal features, but we do not report those results because they did not improve performance over I3D features alone. We believe this is because I3D features already include enough image information for the video description task. In comparison to C3D, which uses the VGG-16 base architecture and was trained on the Sports-1M dataset~\cite{karpathy2014large}, I3D uses a more powerful Inception-V3 network architecture and was trained on the larger (and cleaner) Kinectics~\cite{kay2017kinetics} dataset. As a result, I3D has demonstrated state-of-the-art performance for the task of human action recognition in video sequences~\cite{carreira2017quo}. Further, the Inception-V3 architecture has significantly fewer network parameters than the VGG-16 network, making it more efficient. %As a result, the I3D model can take as input longer sequences (64 frames) versus the 16 frames of C3D---such longer temporal receptive fields lead to better characterization of the spatiotemporal action dynamics.

%VGGish
In terms of audio features, the Audio Set VGGish model provided the best performance.  While we expected the deep features (SoundNet and VGGish) to provide improved performance compared to MFCC, there are several possibilities as to why VGGish performed better than SoundNet.  First, the VGGish model was trained on more data, and had audio specific labels, whereas SoundNet used pre-trained image classification networks to provide labels for training the audio network.  Second, the large Audio Set ontology used to train VGGish likely provides the ability to learn features more relevant to text descriptions than the broad scene/object labels used by SoundNet.

\if 0
%The Na\"ive Fusion model performed better than the Unimodal models.
The results demonstrate the effectiveness of our proposed model. In Table~\ref{table:result1}, the proposed methods outperform the previously published results in all but one evaluation metric of one previous method. In both Tables \ref{table:result1} and~\ref{table:result2}, our proposed methods outperform the ``Na\"ive Fusion (V)'' baseline, which is our implementation of the state-of-the-art methods~\cite{attention4spatial@CVPR2016} and~\cite{MSR-VTT:CVPR16}. 
Furthermore, our proposed Attentional Fusion model outperforms the corresponding
Na\"ive Fusion model, both with audio features (AV) and without audio features (V), 
on both datasets. These results clearly demonstrate the benefits of our proposed multimodal attention model.
%Table \ref{table:sample} shows generated descriptions for four example videos from the YouTube2Text data set. These and more examples, including the original videos with sound, are in the supplementary material.
\fi

\setlength\tabcolsep{2pt} % default value: 6pt
%--------------------------------------------------
\begin{table}[t]
\begin{center}
\caption{Video description evaluation results on the MSVD (YouTube2Text) test set.}
\label{table:result1}

%--------------------------------------------------
\small
\vspace{4pt}
%\begin{tabular}{c||c|c|c|c||c|c|c}
\begin{tabular}{c|c|c||c|c|c}
\multicolumn{6}{c}{{\bf MSVD (YouTube2Text) Full Dataset}}\\
\hline
\multicolumn{3}{c||}{Modalities (feature types)} &    \multicolumn{3}{|c}{Evaluation metric} \\
\hline    
	 %Fusion method & Attention 
	 Image & Spatiotemporal & Audio & BLEU4 & METEOR & CIDEr \\
\hline

%------------------------------
%\parbox{3cm}{\centering {
 VGG-16 & C3D &   
& 0.524 %BLEU4
& 0.320 %METEOR
& 0.688 %CIDEr
\\

%\parbox{3cm}{\centering {
VGG-16 & C3D & MFCC 
& 0.539 %BLEU4
& 0.322 %METEOR
& 0.674  %CIDEr
\\
\hline

\if 0
%------------------------------
%\parbox{3cm}{\centering {
%Na\"ive Fusion (V) 
%\\\hspace{5mm} ( RMSprop )}} 
%& Temporal 
%& VGG-16 & C3D & MFCC
& I3D (rgb-flow) &
&  %BLEU4
&  %METEOR
&  %CIDEr
\\[2pt]
%------------------------------
%\parbox{3cm}{\centering {
%Na\"ive Fusion (AV) 
%\\\hspace{5mm} ( RMSprop )}} 
%& Temporal 
%& VGG-16 & C3D & MFCC
& I3D (rgb-flow) & VGGish
&  %BLEU4
&  %METEOR
&  %CIDEr
\\[2pt]
\fi 

\hline
%------------------------------
%\parbox{3cm}{\centering {
%{Attentional Fusion (V)} 
%\\ ( AdaDelta )}} 
%& {Temporal \& Multimodal}
%& VGG-16 & C3D & 
& I3D (rgb-flow) &  
&  0.525  %BLEU4
&  0.330 %METEOR
&  0.742 %CIDEr
\\
\hline
%------------------------------
%\parbox{3cm}{\centering {
%
%& 
& &  MFCC
& 0.527 %BLEU4
& 0.325 %METEOR
& 0.702  %CIDEr
\\

%------------------------------
%\parbox{3cm}{\centering {
%{Attentional Fusion (AV)} 
%& {Temporal \& Multimodal} 
& I3D (rgb-flow) & SoundNet
&  0.529 %BLEU4
&  0.319 %METEOR
&  0.719 %CIDEr
\\

%------------------------------
%\parbox{3cm}{\centering {
%
%& 
 & & VGGish
& {\bf 0.554} %BLEU4
& {\bf 0.332} %METEOR
& {\bf 0.743} %CIDEr
\\

\hline
\end{tabular}
\end{center}
\vspace{-6pt}
\end{table}
%}
%--------------------------------------------------

%--------------------------------------------------
\begin{table}[t]
\begin{center}
\caption{Video description evaluation results on MSR-VTT Subset. Approximately 12\% of the MSR-VTT videos have been removed from YouTube, so we train and test on the remaining Subset of MSR-VTT videos that we were able to download.
The normalization for the visual features was not applied to MSR-VTT in this experiments.
%
%We cannot directly compare with the results in~\cite{MSR-VTT:CVPR16}, because they used the full MSR-VTT dataset. Our Na\"ive Fusion (V) baseline method is extremely similar to the method of ~\cite{MSR-VTT:CVPR16}, so it may be viewed as our implementation of their method using the available subset of the MSR-VTT dataset.
%115260 sentences
%We split the data according to 65\%:30\%:5\%, 
%corresponding to 6,513, 2,990 and 497 clips
%in the training, testing and validation sets, respectively
}
\label{table:result2}
%--------------------------------------------------
\small
\vspace{4pt}
%\begin{tabular}{c||c|c|c|c||c|c|c}
\begin{tabular}{c|c|c||c|c|c}
\multicolumn{6}{c}{{\bf MSR-VTT Subset}}\\
\hline
\multicolumn{3}{c||}{Modalities (feature types)} &    \multicolumn{3}{|c}{Evaluation metric} \\
	  \hline
	 %Fusion method & Attention 
	 Image & Spatiotemporal & Audio & BLEU4 & METEOR & CIDEr \\
\hline

%\parbox{3cm}{\centering {
%{Attentional Fusion (AV)} 
%\\( AdaDelta )}} 
%& { Temporal \& Multimodal} 
VGG-16 & C3D & MFCC &
{\bf 0.397} & %BLEU4
0.255 & %METEOR
0.400  %CIDEr
\\
\hline

\if 0
%------------------------------
%\parbox{3cm}{\centering {
%Na\"ive Fusion (V) 
%\\\hspace{5mm} ( RMSprop )}} 
%& Temporal 
%& VGG-16 & C3D & MFCC
&I3D (rgb-flow) &
&  %BLEU4
&  %METEOR
&  %CIDEr
\\[2pt]
%------------------------------
%\parbox{3cm}{\centering {
%Na\"ive Fusion (AV) 
%\\\hspace{5mm} ( RMSprop )}} 
%& Temporal 
%& VGG-16 & C3D & MFCC
&I3D (rgb-flow) & VGGish
&  %BLEU4
&  %METEOR
&  %CIDEr
\\[2pt]
\fi 

\hline
%------------------------------
%\parbox{3cm}{\centering {
%{Attentional Fusion (V)} 
%\\ ( AdaDelta )}} 
%& {Temporal \& Multimodal}
%& VGG-16 & C3D & 
&I3D (rgb-flow) &  
&  0.347 %BLEU4
&  0.241 %METEOR
&  0.349 %CIDEr
\\
\hline
%------------------------------
%\parbox{3cm}{\centering {
%
%& 
& &  MFCC
& 0.364 %BLEU4
& 0.253 %METEOR
& 0.393  %CIDEr
\\

%------------------------------
%\parbox{3cm}{\centering {
%{Attentional Fusion (AV)} 
%& {Temporal \& Multimodal} 
& I3D (rgb-flow) & SoundNet
& 0.366 %BLEU4
& 0.246 %METEOR
& 0.387 %CIDEr
\\

%------------------------------
%\parbox{3cm}{\centering {
%
%& 
 & & VGGish
& 0.390 %BLEU4
& {\bf 0.263} %METEOR
& {\bf 0.417} %CIDEr
\\

\hline
\end{tabular}
\end{center}
\vspace{-6pt}
\end{table}
%}
%--------------------------------------------------

%--------------------------------------------------
\begin{table}[htb]
\begin{center}
\caption{Video description evaluation results on Charades. 
%In this experiment, SoundNet was not tested.
}
\label{table:result3}
%--------------------------------------------------
\small
\vspace{4pt}
%\begin{tabular}{c||c|c|c|c||c|c|c}
\begin{tabular}{c|c|c||c|c|c}
\multicolumn{6}{c}{{\bf Charades Dataset}}\\
\hline
\multicolumn{3}{c||}{Modalities (feature types)} &    \multicolumn{3}{|c}{Evaluation metric} \\
	  \hline
	 %Fusion method & Attention 
	 Image & Spatiotemporal & Audio & BLEU4 & METEOR & CIDEr \\
\hline

\if 0
%------------------------------
%\parbox{3cm}{\centering {
%Na\"ive Fusion (V) 
%\\\hspace{5mm} ( RMSprop )}} 
%& Temporal 
%& VGG-16 & C3D & MFCC
& I3D (rgb-flow) &
&  %BLEU4
&  %METEOR
&  %CIDEr
\\[2pt]
%------------------------------
%\parbox{3cm}{\centering {
%Na\"ive Fusion (AV) 
%\\\hspace{5mm} ( RMSprop )}} 
%& Temporal 
%& VGG-16 & C3D & MFCC
& I3D (rgb-flow) & VGGish
&  %BLEU4
&  %METEOR
&  %CIDEr
\\[2pt]
\fi 

\hline
%------------------------------
%\parbox{3cm}{\centering {
%{Attentional Fusion (V)} 
%\\ ( AdaDelta )}} 
%& {Temporal \& Multimodal}
%& VGG-16 & C3D & 
& I3D (rgb-flow) &  
& 0.094 %BLEU4
& 0.149 %METEOR
& 0.236 %CIDEr
\\
\hline
%------------------------------
%\parbox{3cm}{\centering {
%
%& 
& &  MFCC
& 0.098  %BLEU4
& 0.156 %METEOR
& 0.268  %CIDEr
\\

%------------------------------
%\parbox{3cm}{\centering {
%{Attentional Fusion (AV)} 
%& {Temporal \& Multimodal} 
& I3D (rgb-flow) & SoundNet
& - %BLEU4
& - %METEOR
& - %CIDEr
\\

%------------------------------
%\parbox{3cm}{\centering {
%
%& 
 & & VGGish
& {\bf 0.100} %BLEU4
& {\bf 0.157} %METEOR
& {\bf 0.270} %CIDEr
\\

\hline
\end{tabular}
\end{center}
\vspace{-6pt}
\end{table}
\begin{table*}[t]
%\begin{center}
\footnotesize
\centering
\caption{System response generation evaluation results with objective measures.}
\label{tab:one-ref-result}
\begin{tabular}{l|c|ccccccc}
\hline
               & Attentional &   &  & & & & & \\
\raisebox{1.8mm}[0ex]{Input features} & fusion & 
\raisebox{1.8mm}[0ex]{BLEU1} &
\raisebox{1.8mm}[0ex]{BLEU2} &
\raisebox{1.8mm}[0ex]{BLEU3} &
\raisebox{1.8mm}[0ex]{BLEU4} & 
\raisebox{1.8mm}[0ex]{METEOR} &
\raisebox{1.8mm}[0ex]{ROUGE\_L} &
\raisebox{1.8mm}[0ex]{CIDEr} \\
\hline
QA              & - & 0.236 & 0.142 & 0.094 & 0.065 &  0.101 & 0.257 & 0.595  \\
QA + Captions   & - & 0.245 & 0.152 & 0.103 & 0.073 &  0.109 & 0.271 & 0.705  \\
QA + VGG16      & -    & 0.231   & 0.141 & 0.095 & 0.067 & 0.102 & 0.259 & 0.618   \\
QA + I3D        & no   & 0.246  & 0.153 & 0.104 & 0.073 & 0.109 & 0.269 & 0.680   \\
QA + I3D        & yes & 0.250  & 0.157 & 0.108 & 0.077 & 0.110 & 0.274 & 0.724   \\
QA + I3D + VGGish &  no        & 0.249 & 0.155 & 0.106 & 0.075 & 0.110 & 0.275 & 0.701 \\
QA + I3D + VGGish & yes & {\bf 0.256} & {\bf 0.161} & {\bf 0.109} & {\bf 0.078} & {\bf 0.113} & {\bf 0.277} & {\bf 0.727} \\
\hline
\end{tabular}
%\end{center}
%\vskip -3mm
\end{table*}
%----------------------------------------------------------
%\fi 

Since it is intractable to enumerate all possible word sequences in vocabulary $\cal V$, we usually limit them to the $n$-best hypotheses generated by the system. Although in theory the distribution $P(Y'|X)$ should be the true distribution, we instead estimate it using the encoder-decoder model.

\if 0
%%-------------------------
\begin{table}[t]
\centering
\caption{Model size}
\label{tab:model}
\begin{tabular}{l|ccc|cc}
\hline
       & \multicolumn{3}{|c}{encoder} & \multicolumn{2}{|c}{decoder} \\
       \cline{2-6}
       & \#layer & \#sent-layer & \#cell & \#layer & \#cell \\
        \hline
LSTM  & 2 & - & 128 & 2 & 128 \\
BLSTM & 2 & - & 128 & 2 & 256 \\
HRED  & 2 & 1 & 128 & 2 & 128 \\
\hline
\end{tabular}
\vskip -5mm
\end{table}
%----------------------------
\fi

%%%%%%%%%%%%%%%%%%%%%%%%%%
% End of Page 6
%%%%%%%%%%%%%%%%%%%%%%%%%%
%\fi
%\if 0

%%%%%%%%%%%%%%%%%%%%%%%%%%%%%%%%%%%%%%%%
\section{Experiments for Video-scene-aware Dialog}
%%%%%%%%%%%%%%%%%%%%%%%%%%%%%%%%%%%%%%%%
\label{sec:experiments}

%End-to-End dialog
In this paper, we extended an end-to-end dialog system to scene-aware dialog with multimodal fusion.
As shown in Fig.~\ref{fig:scene-aware-dialog}, we embed the video and audio features selected in Section \ref{sec:video-features}.

\if 0
% Our proposal for DSTC6, what's new?
Our proposed system has several key features including
sequence adversarial training, example-based response selection, multiple sequence-to-sequence models, and minimum Bayes risk (MBR) decoding, where
the multiple models are a long short-term memory (LSTM) encoder decoder, a bidirectional LSTM (BLSTM) encoder decoder, and a hierarchical recurrent encoder decoder (HRED). 
A system combination is performed to combine the multiple hypotheses from these models to improve BLEU score.
Sequence adversarial training and the example-based method are used to obtain a high human rating score. 
Experimental results on the Twitter help-desk dialog task show that adversarial training and the example-based method are effective in improving human rating score while system combination improves objective measures such as BLEU and METEOR scores (We might change this part after getting human rating results).
Furthermore, we investigate extension of reward functions for sequence adversarial training to balance subjective and objective scores.
\fi

\subsection{Conditions}
We evaluated our proposed system with the dialog data for Charades we collected.
Table \ref{tab:data} shows the size of each data set. 
We compared the performance between models trained from various combinations of the QA text, visual and audio features. 
In addition, we tested an efficacy of multimodal-attention mechanism for dialg response generation.
\if 0
In order to be able to predict responses occurring partway through a dialog, 
we expanded the training and development sets by truncating each dialog after each system response, and adding the truncated dialogs to the data sets.
% duplicating incomplete dialogs as individual dialogs.
In each dialog, all turns except the last response were concatenated into one sequence to form input sequence $X$, with meta symbols {\tt <U>} and {\tt <S>} inserted at the beginning of each turn to explicitly utilize turn switching information. The last response was used as output sequence $Y$.

We built a LSTM model for response generation using the expanded training set. \fi 
We employed an ADAM optimizer \cite{kingma2014adam} with the cross-entropy criterion and iterated the training process up to 20 epochs. For each of the encoder-decoder model types, we selected the model with the lowest perplexity on the expanded development set. %We also decided the model size based on the BLEU score for the development set.
%, which resulted in Table \ref{tab:model}.

We used the parameters of the LSTMs with \#layer=2 and \#cells=128 for encoding history and question sentences. Video features were projected to 256 dimensional feature space before modality fusion. The decoder LSTM had a structure of \#layer=2 and \#cells=128 as well.

\subsection{Evaluation Results}
Table \ref{tab:one-ref-result} shows the response sentence generation performance of our models, training and decoding methods using objective measures, BLEU1-4, METEOR, ROUGE\_L, and CIDEr, which were computed with the evaluation code for MS COCO caption generation as done for video description.
We investigated different input features including question-answering dialog history plus last question (QA), human-annotated captions (Captions), video features of VGG16 or I3D rgb and flow features (I3D), and audio features (VGGish).

First we evaluated response generation quality with only QA features as a baseline without any video scene features.
Then, we added the caption features to QA, and the performance improved significantly. This is because each caption provided the scene information in natural language and helped the system answer the question correctly.
However, such human annotations are not available for real systems.

Next we added VGG16 features to QA, but they did not increase the evaluation scores from those of QA-only features. This result indicates that QA+VGG16 is not enough to let the system generate better responses than those of QA+Captions.
After that, we replaced VGG16 with I3D, and obtained a certain improvement from the QA-only case. As in the video description, it has been shown that the I3D features are also useful for scene-aware dialog. Furthermore, we applied the multi-modal attention mechanism (attentional fusion) for I3D rgb and flow features, and obtained further improvement in all the metrics.

Finally, we examined the efficacy of audio features. The table shows that VGGish obviously contributed to increasing the response quality especially when using the attentional fusion.
The following example of system response was obtained with or without VGGish features, which worked better for the questions regarding audios:

%%%%%%%%%%%%%%%%%%%%%%%%
%Sample for Page 8
%\if 0
%%%%%%%%%%%%%%%%%%%%%%%%
\begin{tabular}{|ll|}
\hline
\texttt{Question:}      & \texttt{was there audio ? }\\
\texttt{Ground truth:} &\texttt{there is audio , i can hear music and background noise . }\\
\texttt{I3D:}          & \texttt{no , there is no sound in the video . }\\
\texttt{I3D+VGGish:}   & \texttt{yes there is sound in the video .}\\ 
\hline
\end{tabular}
%\fi 

%%%%%%%%%%%%%%%%%%%%%%%%%%%%%%%%%%%%%
\section{Conclusion}
%%%%%%%%%%%%%%%%%%%%%%%%%%%%%%%%%%%%%
In this paper, we propose a new research target, a dialog system that can discuss dynamic scenes with humans, which lies at the intersection of multiple avenues of research in natural language processing, computer vision, and audio processing. 
To advance this goal, we introduce a new model that incorporates technologies for multimodal attention-based video description into an end-to-end dialog system. 
We also introduce a new dataset of human dialogues about videos.  
%We built an end-to-end dialog system 
Using this new dataset, we trained an end-to-end conversation model that generates system responses in a dialog about an input video. 
Our experiments demonstrate that using multimodal features that were developed for multimodal attention-based video description enhances the quality of generated dialog about dynamic scenes.
We are making our data set and model publicly available for a new Video Scene-Aware Dialog challenge.

%\fi
%%%%%%%%%%%%%%%%
% Page 8 to 9
%%%%%%%%%%%%%%%%
%\fi

%\if 0
\bibliographystyle{IEEEbib}
\bibliography{DSTC7-proposal,Sound&Sight,Mitsubishi@DSTC6}
%\fi

\end{document}